\pdfoutput=1

\documentclass[11pt]{article}

\usepackage[final]{acl}

\usepackage{times}
\usepackage{latexsym}

\usepackage[T1]{fontenc}

\usepackage{microtype}

\usepackage{inconsolata}

\usepackage{graphicx}

\usepackage{multirow}

\usepackage{latexsym}
\usepackage{amssymb}
\usepackage{amsmath}
\usepackage{amsthm}
\usepackage{booktabs}
\usepackage{enumitem}
\usepackage{graphicx}
\usepackage{color}
\usepackage{comment}
\usepackage{times}
\usepackage{soul}
\usepackage{url}
\usepackage{hyperref}
\usepackage[utf8]{inputenc}
\usepackage{caption}
\usepackage{algorithm}
\usepackage{algorithmic}
\usepackage{placeins}
\usepackage[switch]{lineno}
\usepackage{subfig}
\usepackage{amsfonts}
\usepackage{xspace}
\usepackage{caption}
\usepackage{diagbox}
\usepackage{makecell}
\usepackage{tablefootnote}
\usepackage{xcolor}
\usepackage[normalem]{ulem}

\newcommand{\cnndm}[0]{\textsc{CNN-DM}\xspace}
\newcommand{\humaneval}[0]{\textsc{Human-Eval}\xspace}
\newcommand{\gsm}[0]{\textsc{GSM-8k}\xspace}
\newcommand{\red}[1]{\textcolor{red}{#1}}
\newcommand{\blue}[1]{\textcolor{blue}{#1}}
\newcommand{\llama}[0]{\textsc{Llama}\xspace}
\newcommand{\vicuna}[0]{\textsc{Vicuna}\xspace}
\newcommand{\flan}[0]{\textsc{Flan-T5}\xspace}
\newcommand{\qwen}[0]{\textsc{Qwen3}\xspace}
\newcommand{\medusa}[0]{\textsc{Medusa}\xspace}
\newcommand{\const}[0]{\textsc{CoGA}\xspace}
\newcommand{\heuristic}[0]{\textsc{DyGA}\xspace}
\newcommand{\eagle}[0]{\textsc{Eagle}\xspace}
\newcommand{\vanilla}[0]{\textsc{Vanilla}\xspace}
\newcommand{\vv}[0]{\textsc{V}\xspace}

\newcommand{\qw}[0]{\textsc{Q}\xspace}

\newcommand{\redunderline}[1]{\textcolor{red}{\uline{#1}}}

\definecolor{deepgreen}{RGB}{0,90,40}

\newcommand{\greenunderline}[1]{\textcolor{deepgreen}{\uline{#1}}}

\newcolumntype{P}[1]{>{\centering\arraybackslash}m{#1}}

\title{Benchmarking the Energy Savings with Speculative Decoding Strategies}

\author{
\textbf{Rohit Dutta\textsuperscript{1}},
\textbf{Paramita Koley\textsuperscript{2}},
 \textbf{Soham Poddar\textsuperscript{1}},
 \textbf{Janardan Misra\textsuperscript{3}},\\
 \textbf{Sanjay Podder\textsuperscript{3}},
 \textbf{Naveen Balani\textsuperscript{3}},
 \textbf{Saptarshi Ghosh\textsuperscript{1},
 \textbf{Niloy Ganguly\textsuperscript{1}}} \\
 \textsuperscript{1} Indian Institute of Technology, Kharagpur, India \\
 \textsuperscript{2} Indian Statistical Institute, Kolkata, India \\
 \textsuperscript{3} Accenture Labs, Bangalore, India \\
}

\begin{document}
\maketitle

\begin{abstract}

Speculative decoding has emerged as an effective method to reduce latency and inference cost of LLM inferences. However, there has been inadequate attention towards the energy requirements of these models. To address this gap, this paper presents a comprehensive survey of energy requirements of speculative decoding strategies, with detailed analysis on how various factors -- model size and family, speculative decoding strategies, and dataset characteristics -- influence the energy optimizations.

\end{abstract} 

\section{Introduction}

Large Language Models (LLMs) have witnessed rapid adoption across a wide range of applications. Despite their utility, the deployment of these models demands substantial computational resources, leading to considerable energy consumption~\cite{wu2022sustainable, patterson2022carbon,poddar2025benchmarking}. Recent work by \citet{poddar2025benchmarking} identifies token decoding latency and model complexity as critical determinants of inference-time energy consumption. While autoregressive models, forming the basis of most LLMs, generate tokens in a sequential manner, inherently resulting in higher decoding latency, \textit{speculative decoding}~\cite{leviathan2023fast, chen2023accelerating} has emerged as a promising approach to reduce decoding time. This approach leverages a \textit{lightweight `assistant model'} to generate candidate token sequences, followed by a parallelized verification stage in which the larger `\textit{target model}' evaluates multiple tokens in a single pass. This mechanism significantly reduces decoding time while offloading a substantial portion of the sequential generation to a smaller, more efficient model. Given these characteristics, we hypothesize that speculative decoding may also offer reductions in inference-time energy consumption. 

While the existing literature on speculative decoding ~\cite{leviathan2023fast, cai2024medusa, eagle2, eagle3} has predominantly focused on optimizing response time or latency, its implications for energy efficiency remain largely unexplored. To address this gap, this paper presents a comprehensive survey of speculative decoding techniques, with emphasis on their implications for energy efficiency during inference.

\begin{figure}[t]
    \centering
    \subfloat{\includegraphics[width=0.45\textwidth]{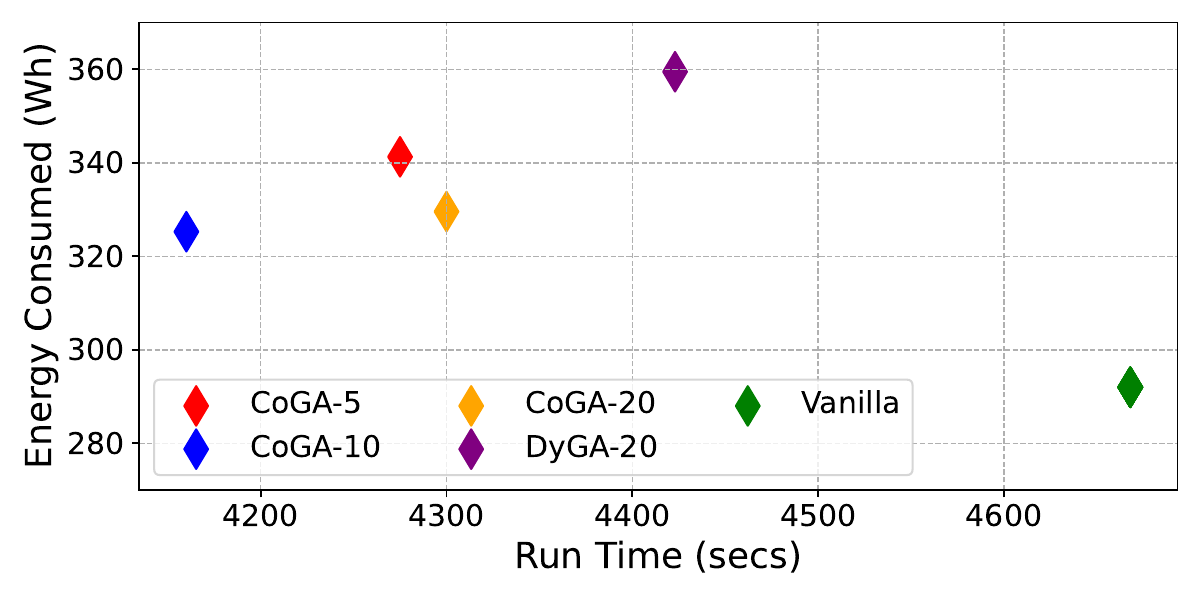}}
    \vspace{-2mm}
    \caption{Speculative decoding approaches (\const, \heuristic) with \vicuna-13B as target ends up in consuming more energy than vanilla decoding (applying only target model) despite exhibiting lower inference latency.}
    \vspace{-6mm}
    \label{fig:motivating-example}
\end{figure}

\vspace{1mm}
\noindent \textbf{Motivation:}
Given that speculative decoding (SD) strategies yield speedup in inference latency/time, and response time is strongly correlated with energy consumption~\cite{poddar2025benchmarking}, one might assume that speculative decoding inherently leads to improved energy efficiency. 
However, we argue that walltime speedup~\cite{eagle2, eagle3} 
does \textit{not} necessarily translate into proportional energy savings. Furthermore, even when such a correlation exists, the relationship between runtime and energy consumption is often nonlinear and influenced by multiple interacting factors.

To substantiate this claim, we conducted a preliminary experiment utilizing two representative speculative decoding methods -- \const~\cite{leviathan2023fast} and  heuristic~\cite{huggingface2023transformers} -- with \vicuna-13B~\cite{chiang2023vicuna} as the target model and \vicuna-68M as the assistant model (details in later sections). Figure~\ref{fig:motivating-example} reports the runtime versus energy consumption. It illustrates that \textit{the vanilla decoding (i.e., target model alone) consumes less energy than the standard speculative decoding approaches, despite having longer runtime.} This counterintuitive observation leads to a crucial insight: improvements in inference latency do not inherently yield corresponding reductions in energy consumption, emphasizing a clear need for a systematic analysis of SD strategies from the standpoint of energy efficiency. 

In this study, we examine a diverse set of SD techniques across multiple architectures and benchmark datasets, toward elucidating the primary factors governing energy consumption in these approaches. We find that lower inference time may not always translate into proportional energy savings,
and that model family and size difference have crucial roles in energy savings.

\section{Experimental Setup}

\noindent \textbf{Tasks and Datasets:}
We conduct experiments on three standard inference tasks, common in speculative decoding~\cite{leviathan2023fast, eagle2, eagle3}, namely 
(1)~code generation (\humaneval~\cite{humaneval}), 
(2)~mathematical reasoning (\gsm~\cite{gsm8k}), and 
(3)~summarization (\cnndm~\cite{cnndm}). We use  $256$ randomly sampled prompts from each dataset/task except \humaneval which has only $164$ samples. We employed model-specific chat templates; detailed prompts for the three tasks are illustrated in Table~\ref{table:dataset-gsm8k}, Table~\ref{table:dataset-humaneval}, and Table~\ref{table:dataset-cnndm} in Appendix~\ref{app:datasets}). 

\noindent \textbf{Target and Assistant Models:}
We consider LLMs from four families. Specifically, we evaluate: (i)~\vicuna-7B and (ii)~\vicuna-13B as target models paired with \vicuna-68M as the assistant model; (iii)~\llama-8B and (iv)~\llama-70B as target models paired with \llama-1B as the assistant model; (v)~\flan-L and (vi)~\flan-XL as target models paired with \flan-B as the assistant model; and (vii)~\qwen-4B and (viii)~\qwen-8B as target models paired with \qwen-0.6B as the assistant model. All target–assistant model combinations, are listed in the first column of Table~\ref{tab:speed-energy-all}. Across all configurations, the target models have been loaded using NF4 4-bit quantization~\cite{dettmers2023qlora}.
Refer to Table~\ref{tab:model-config} in Appendix~\ref{app:model-details} for model details.

%\subsection{Speculative Decoding strategies}

\noindent \textbf{Speculative Decoding strategies:}
We considered the following SD strategies, namely two variants of standard speculative decoding~\cite{leviathan2023fast} --  
(1)~Constant Generation by Assistant (\const-$x$) and 
(2)~Dynamic Generation by Assistant (\heuristic-$x$) -- 
and two state-of-the-art SD strategies, namely (3)~\eagle-2~\cite{eagle2}, and  (4)~\eagle-3~\cite{eagle3}. We used model-specific chat-templates with greedy decoding for reproducibility.

In \const-$x$, the assistant generates fixed-length drafts ($x$ tokens) per iteration, whereas in \heuristic-$x$, the assistant generates dynamic length drafts, starting with $x$ tokens and adjusting its length in each iteration (by increasing the length by $2$ if all tokens in the last draft is accepted, otherwise reducing the draft length by $1$). For \const-$x$, the value of $x$ was set to $5, 10$ and $20$ in our experiments.

\medusa~\cite{cai2023medusa} mitigates autoregressive bottlenecks by augmenting the target model with multiple decoding heads that predict future tokens in parallel, verified through a tree-based attention mechanism. \eagle-2~\cite{eagle2} builds on standard SD by introducing a context-aware dynamic draft tree for generating dynamic drafts. \eagle-3~\cite{eagle3} further introduces direct token prediction, allowing the draft model to integrate multi-layer fused features from the target model.
Refer Appendix~\ref{app:specdec-strat} for further details. 

For the purpose of  inferences, we employ HugginFace implementations of models~\cite{wolf2019huggingface, huggingface2023transformers}.

\noindent \textbf{Hardware and Energy metrics:}
Most experiments are performed on a single NVIDIA A5000 GPU with 24GB VRAM hosted in a local server with Intel Xeon Silver 4210R processor and 128GB RAM, running Ubuntu 20.04-LTS with Pytorch v2.6 (with CUDA 12.8) and Huggingface transformers v4.51. The experiments comprising 70B models are performed on a single NVIDIA A6000 GPU with 48GB VRAM and other identical configurations as stated above.

We use the popular Code Carbon~\cite{schmidt2021codecarbon} package to measure the energy consumed. During inference, we provide test samples sequentially at batch size $1$ to the LLM. Refer to Appendix~\ref{app:hardware} for further details. 

\vspace{1mm}
\noindent \textbf{Evaluation Metrics for SD performance:} We employ the following metrics for measuring energy and time performance of SD strategies:

\vspace{1mm}
\noindent \textbf{(i) GPU energy saving factor ($\gamma_{e}^{GPU}$)} denotes the ratio of GPU energy consumed by the  target model under vanilla autoregressive decoding and GPU energy consumed under the speculative decoding (SD) strategy.
$\mathbf{\gamma_{e}^{GPU} = \text{Energy}_\text{Target}^\text{GPU}
\; /  \; \text{Energy}_\text{SD}^\text{GPU}}$,
$SD \in \{\const, \heuristic, \eagle, \medusa\}$.

\noindent \textbf{(ii) Total energy saving factor ($\gamma_{e}^{Total}$)} denotes the ratio between the total energy consumed by the target model under vanilla autoregressive decoding and the total energy consumed under the SD strategy. In this context, total energy refers to the cumulative energy drawn by the GPU, CPU, and RAM.
$\mathbf{\gamma_{e}^{Total} = \text{Energy}_\text{Target}^\text{Total}
\; /  \; \text{Energy}_\text{SD}^\text{Total}}$, $\text{SD} \in \{\text{\const}, \text{\heuristic}, \text{\eagle}, \text{\medusa}\}$.

\noindent For both energy gain metrics, we consider the average energy consumed to generate $1K$ tokens.% for the same task.  

\vspace{1mm}
\noindent \textbf{(iii) Speedup ($\gamma_t$)} denotes the ratio of inference time under vanilla autoregressive decoding of the target model and the inference time under speculative decoding (SD) strategy.  
$\gamma_t = \text{Time}_\text{Target} \; / \;  
        \text{Time}_\text{SD}$,
        $\text{SD} \in \{\text{\const}, \text{\heuristic}, \text{\eagle}, \text{\medusa}\}$.
For both cases, we consider the average time to decode $1K$ tokens for the same task.

\noindent Refer to Appendix~\ref{app:metrics} for additional metrics (reported in Table~\ref{table:specdec-energy-comp}).

\section{Energy Consumption of SD Strategies}

In this section, we analyze the energy consumption of SD strategies to identify the factors driving energy efficiency and optimization.
Table~\ref{tab:speed-energy-all} reports the speedup and energy saving factor relative to the vanilla autoregressive decoding setting (where only the target model is run for the whole task) for all datasets and  for all the settings described earlier. We observe that the simpler SD strategies \const and \heuristic achieve energy reduction primarily for \llama and \flan family (up to $2.0 \times$), on all datasets except \cnndm. 
But SOTA SD strategies  \eagle-2 and \eagle-3 achieve notable energy reduction ($1.34 \times$ - $2.51 \times$) on all four model settings across all datasets except (\llama-70B, \cnndm) model-dataset pair. An explanation can be that high speedup in \eagle methods results in notable energy saving. For \const and \heuristic, relatively smaller speedup fails to reflect into useful energy saving.

\begin{table*}[tb]
    \scriptsize
    \centering
        \begin{tabular}{
        p{17mm} p{11mm}| 
        p{9mm} p{10mm} p{10mm}| 
        p{9mm} p{9mm} p{9mm}|
        p{9mm} p{9mm} p{9mm}}
        \toprule
        &  
        & \multicolumn{3}{c|}{\textbf{\humaneval}} 
        & \multicolumn{3}{c|}{\textbf{\gsm}} 
        & \multicolumn{3}{c}{\textbf{\cnndm}}         
        \\        
        \midrule
        & &
        \multicolumn{1}{c}{\textbf{Speedup}} &
        \multicolumn{2}{c|}{\textbf{Energy Saving Factor}} &
        \multicolumn{1}{c}{\textbf{Speedup}} &
        \multicolumn{2}{c|}{\textbf{Energy Saving Factor}} &
        \multicolumn{1}{c}{\textbf{Speedup}} &
        \multicolumn{2}{c}{\textbf{Energy Saving Factor}} \\
        \cmidrule(lr){3-3} \cmidrule(lr){4-5}
        \cmidrule(lr){6-6} \cmidrule(lr){7-8}
        \cmidrule(lr){9-9} \cmidrule(lr){10-11}
        
        \makecell{\textbf{Target vs}\\\textbf{Assistant}} &
        \makecell{\textbf{SD}\\\textbf{Method}} &
        $\boldsymbol{\displaystyle\gamma_t}$ &
        \makecell{$\boldsymbol{\displaystyle\gamma_e^{GPU}}$} &
        \makecell{$\boldsymbol{\displaystyle\gamma_e^{Total}}$} &
        $\boldsymbol{\displaystyle\gamma_t}$ &
        \makecell{$\boldsymbol{\displaystyle\gamma_e^{GPU}}$} &
        \makecell{$\boldsymbol{\displaystyle\gamma_e^{Total}}$} &
        $\boldsymbol{\displaystyle\gamma_t}$ &
        \makecell{$\boldsymbol{\displaystyle\gamma_e^{GPU}}$} &
        \makecell{$\boldsymbol{\displaystyle\gamma_e^{Total}}$} 
        \\
        \midrule
        \multirow{4}{*}{\makecell[l]{\vicuna-7B vs \\ \vicuna-68M}}  
        & \const-20 & $1.42\times$ & $0.83\times$ & $0.99\times$ & $1.19\times$ & $0.72\times$ & $0.77\times$ & $1.44\times$ & $0.89\times$ & $1.03\times$ \\
        & \heuristic-20 & $1.43\times$ & $0.8\times$ & $0.96\times$ & $1.28\times$ & $0.73\times$ & $0.84\times$ & $1.34\times$ & $0.82\times$ & $0.94\times$ \\
        & \eagle2 & \greenunderline{$2.86\times$} & $1.57\times$ & \greenunderline{$1.90\times$} & \greenunderline{$2.47\times$} & $1.37\times$ & $1.61\times$ & \greenunderline{$1.98\times$} & $1.17\times$ & \greenunderline{$1.38\times$}\\
        & \medusa & $1.87\times$ & \greenunderline{$1.78\times$} & $1.76\times$ & $2.09\times$ & \greenunderline{$1.85\times$} & \greenunderline{$1.90\times$} & $1.26\times$ & \greenunderline{$1.21\times$} & $1.20\times$\\
        \midrule
        \multirow{5}{*}{\makecell[l]{\vicuna-13B vs \\ \vicuna-68M}}
        & \const-20 & $1.07\times$ & $0.93\times$ & $0.92\times$ & $1.08\times$ & $0.75\times$ & $0.88\times$ & $1.02\times$ & $0.89\times$ & $0.96\times$ \\
        & \heuristic-20 & $1.05\times$ & $0.88\times$ & $0.92\times$ & $1.05\times$ & $0.71\times$ & $0.81\times$ & $0.98\times$ & $0.85\times$ & $0.91\times$ \\
        & \eagle2 & $2.16\times$ & $1.79\times$ & $1.83\times$ & $2.15\times$ & $1.46\times$ & $1.63\times$ & $1.52\times$ & $1.29\times$ & $1.47\times$\\
        & \eagle3 & \greenunderline{$2.91\times$} & \greenunderline{$2.40\times$} & \greenunderline{$2.51\times$} & \greenunderline{$2.76\times$} & $1.87\times$ & \greenunderline{$2.09\times$} & \greenunderline{$2.17\times$} & \greenunderline{$1.86\times$} & \greenunderline{$2.10\times$}\\
        & \medusa & $2.24\times$ & $2.09\times$ & $2.11\times$ & $2.10\times$ & \greenunderline{$2.05\times$} & $2.03\times$ & $1.47\times$ & $1.46\times$ & $1.43\times$\\
        \midrule
        \multirow{4}{*}{\makecell[l]{\llama-8B vs \\ \llama-1B}} 
        & \const-20 & $1.78\times$ & $1.23\times$ & $1.42\times$ & $1.59\times$ & $1.10\times$ & $1.19\times$ & $0.96\times$ & $0.79\times$ & $0.83\times$ \\
        & \heuristic-20 & $1.79\times$ & $1.26\times$ & $1.44\times$ & $1.50\times$ & $1.06\times$ & $1.11\times$ & $0.99\times$ & $0.80\times$ & $0.85\times$ \\
        & \eagle2 & $2.30\times$ & $1.35\times$ & $1.63\times$ & $2.01\times$ & $1.19\times$ & $1.34\times$ & $1.48\times$ & $1.03\times$ & $1.21\times$ \\
        & \eagle3 & \greenunderline{$2.90\times$} & \greenunderline{$1.74\times$} & \greenunderline{$2.09\times$} & \greenunderline{$2.62\times$} & \greenunderline{$1.58\times$} & \greenunderline{$1.73\times$} & \greenunderline{$1.84\times$} & \greenunderline{$1.31\times$} & \greenunderline{$1.52\times$}\\
        \midrule
        \multirow{3}{*}{\makecell[l]{\llama-70B vs \\ \llama-1B}} 
        & \const-20 & $1.25\times$ & $1.33\times$ & $1.31\times$ & $1.09\times$ & $1.12\times$ & $1.14\times$ & $0.61\times$ & $0.64\times$ & $0.63\times$ \\
        & \heuristic-20 & $1.29\times$ & \greenunderline{$1.38\times$} & \greenunderline{$1.36\times$} & $1.10\times$ & $1.14\times$ & $1.15\times$ & $0.62\times$ & $0.64\times$ & $0.64\times$ \\
        & \eagle3 & \greenunderline{$1.35\times$} & $1.34\times$ & $1.34\times$ & \greenunderline{$1.28\times$} & \greenunderline{$1.26\times$} & \greenunderline{$1.28\times$} & \redunderline{$0.68\times$} & \redunderline{$0.78\times$} & \redunderline{$0.77\times$}\\
        \midrule
        % \multirow{2}{*}{\makecell[l]{\fl-L vs \fl-B}}
        \flan-L vs & \const-20 & \greenunderline{$2.01\times$} & \greenunderline{$2.02\times$} & \greenunderline{$2.00\times$} & $1.22\times$ & $1.22\times$ & $1.22\times$ & $1.47\times$ & $1.39\times$ & $1.40\times$\\
        \flan-B & \heuristic-20 & $1.97\times$ & $1.95\times$ & $1.94\times$ & \greenunderline{$1.30\times$} & \greenunderline{$1.29\times$} & \greenunderline{$1.29\times$} & \greenunderline{$1.51\times$} & \greenunderline{$1.40\times$} & \greenunderline{$1.42\times$}\\
        \midrule
        % \multirow{2}{*}{\makecell[l]{\fl-XL vs \fl-B}}
        \flan-XL vs & \const-20 & $1.86\times$ & $1.82\times$ & $1.81\times$ & $0.92\times$ & $0.92\times$ & $0.92\times$ & \greenunderline{$1.69\times$} & \greenunderline{$1.45\times$} & \greenunderline{$1.45\times$}\\
        \flan-B & \heuristic-20 & \greenunderline{$1.86\times$} & \greenunderline{$1.85\times$} & \greenunderline{$1.86\times$} & \greenunderline{$1.03\times$} & \greenunderline{$1.00\times$} & \greenunderline{$1.00\times$} & $1.44\times$ & $1.40\times$ & $1.41\times$\\
        \midrule
        % \multirow{3}{*}{\makecell[l]{\qw-4B vs \qw-0.6B}}
        % & \\
        \qw-4B vs \qw-0.6B & \heuristic-20 & \greenunderline{$1.10\times$} & \greenunderline{$1.05\times$} & \greenunderline{$1.05\times$} & \greenunderline{$1.23\times$} & \greenunderline{$1.14\times$} & \greenunderline{$1.15\times$} & \redunderline{$0.93\times$} & \redunderline{$0.84\times$} & \redunderline{$0.85\times$}\\
        % & \\
        \midrule
        % \multirow{3}{*}{\makecell[l]{\qwen-8B vs \\ \qwen-0.6B}}
        % & \\
        \qw-8B vs \qw-0.6B & \heuristic-20 & \greenunderline{$1.09\times$} & \redunderline{$0.79\times$} & \redunderline{$0.80\times$} & \greenunderline{$1.19\times$} & \redunderline{$0.90\times$} & \redunderline{$0.91\times$} & \redunderline{$0.91\times$} & \redunderline{$0.66\times$} & \redunderline{$0.67\times$} \\
        % & \\
        \bottomrule
    \end{tabular}
    \vspace{-2mm}
    \caption{Comparative analysis of speedup ($\gamma_t$), GPU ($\displaystyle\gamma_e^{GPU}$) and Total ($\displaystyle\gamma_e^{Total}$) energy saving factor for various SD strategies. Here, \eagle2 and \eagle3 employ draft models provided by the source repositories. $\gamma_e^{GPU/Total} \ge 1.0\times$ and $\gamma_t \ge 1.0\times$ indicate reduction in energy and time respectively, relative to vanilla decoding.\tablefootnote{The results for \qwen-4B and \qwen-8B are reported using only \heuristic-20 as an implementation of \const-$x$ is not available for \qwen models in the HuggingFace framework.}
    For each dataset–model combination, the best-performing value is highlighted using \protect\greenunderline{green}. In cases where performance degradation persists even under the best configuration, the corresponding metrics are indicated using \protect\redunderline{red}.}
    \vspace{-4mm}
    \label{tab:speed-energy-all}
\end{table*}

\vspace{1mm}
\noindent \textbf{Model-specific trends:} Model family plays a crucial role in energy reduction. Among decoder-only models, \llama models generally achieve moderate to high energy savings (up to $2.09\times$ for \llama-8B and up to $1.36\times$ for \llama-70B), \vicuna models achieve energy savings only on \eagle-(2,3), with no reduction for \const and \heuristic. Among encoder-decoder family, \flan models achieve high energy savings (up to $2.02\times$ for \flan-L and upto $1.86\times$ for \flan-XL). On the other hand, energy savings is minimal for \qwen family ($1.15\times$ for \qwen-4B and $0.91\times$ for \qwen-8B), with no reduction for many cases.
\vspace{1mm}
\noindent \textbf{Dataset-specific trends:} 
Energy reduction substansially varies across datasets, with maximum for \humaneval and minimum for \cnndm. 
\const and \heuristic achieve maximum energy saving of $2.0\times$ on \humaneval with \flan-L. 
\eagle performed the best both latency and energy-wise - achieving highest energy saving of $2.5\times$ times on \humaneval with \vicuna-13B model. 
Surprisingly, \llama-70B, that achieves notable energy saving in most other SD setups, ends up with increased energy consumption with SD in case of the \cnndm dataset (relative to vannila decoding). Similarly, \qwen-4B that achieves $1.05\times$ times energy gain in \humaneval, drops to $0.85\times$ in \cnndm. Thus, energy optimization may vary significantly depending on the task/dataset.

\begin{figure}[tb]
    \centering
    \begin{minipage}{\linewidth}
        {\includegraphics[width=0.9\linewidth]{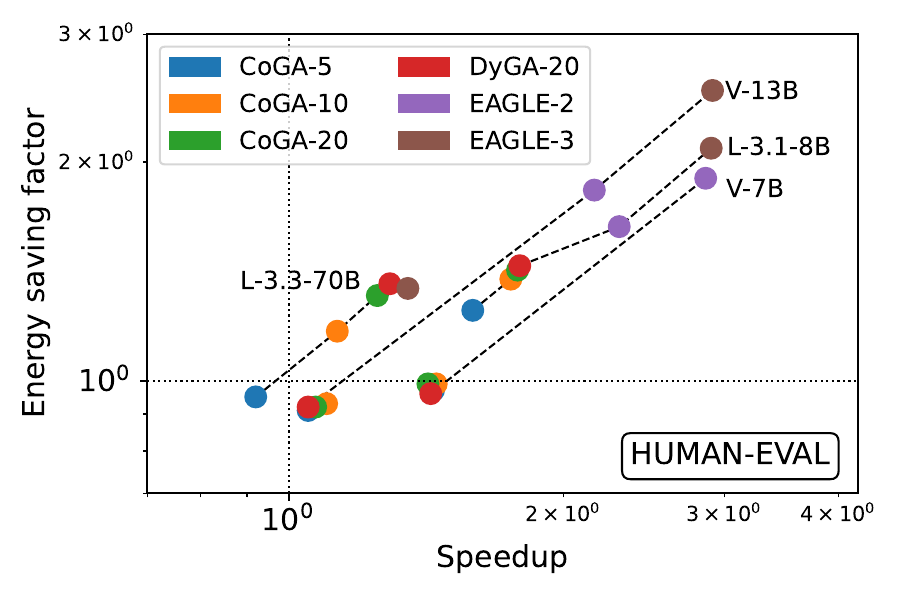}\label{fig:speed-vs-energy-humaneval}} 
        \vspace{-4 mm}
        {\includegraphics[width=0.9\linewidth]{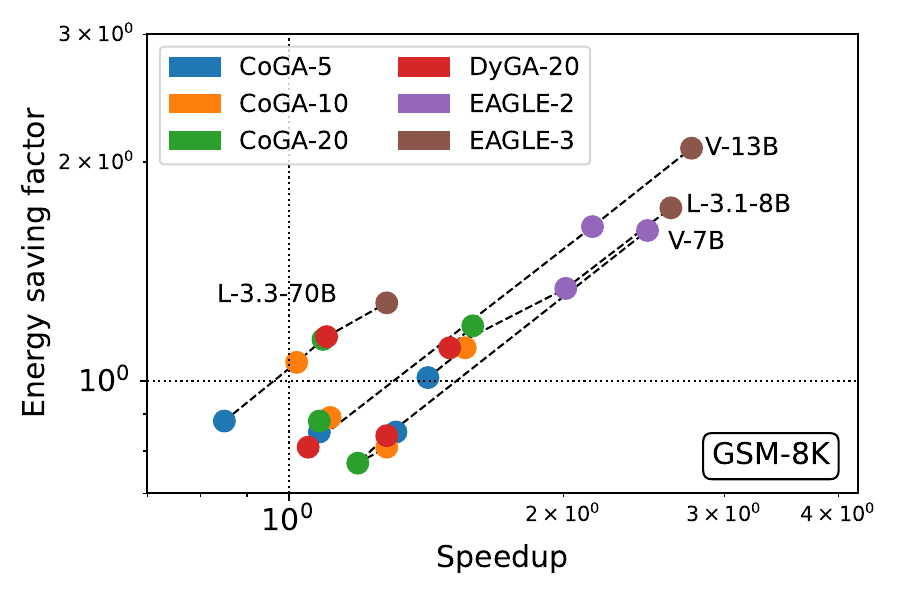}\label{fig:speed-vs-energy-gsm}}
    \end{minipage}
    \caption{Speedup ($\gamma_t$) and total energy savings ($\gamma_e^{\text{Total}}$) for speculative decoding methods across model pairs on \humaneval and \gsm. \eagle consistently achieves the highest speedup and energy efficiency. \llama-based pairs under both \const and \heuristic exhibit positive runtime and energy gains, while \vicuna-based pairs frequently underperform, with values often below unity.}
    \vspace{-5 mm}
\label{fig:speedup-vs-energy}
\end{figure}

\subsection{Analysis of Speedup and Energy Savings}

Figure~\ref{fig:speedup-vs-energy} shows speedup ($\gamma_t$) vs energy saving factor ($\gamma_e^{\text{Total}}$) for all models and SD strategies for the \humaneval and \gsm datasets. A value greater than $1.0$ indicates energy/time savings.
In general, we observe that energy saving factor varies more or less linearly with walltime speedup factor, with the slope varying with the target model chosen. We also observe that larger difference in target and assistant model size results in higher energy optimization, resulting in a steeper slope.
Closer proximity of the lines corresponding to \llama-8B and \vicuna-7B further validates this claim. 
However, this trend is somewhat less prominent if the larger target model is \llama-70B due to the \llama-1B assistant model's outputs not aligning perfectly. Also, for \llama-8B, \eagle achieves higher energy saving relative to speedup, while comparing with \const, and \heuristic, showing that even for same model, the slope can vary across speculative decoding approaches.
\begin{table}[!htb]
    \tiny
    \centering
    \begin{tabular}{P{13mm} P{9mm} P{7mm} P{6mm} P{6mm} P{7mm}}
        \toprule
        \textbf{Target vs Assistant} & \textbf{Method} & \textbf{Assistant Time (mins)} & \textbf{Target Time (mins)} & \textbf{Total Time (mins)} & \textbf{Total Energy (Wh)} \\
        \midrule
        \multicolumn{6}{c}{\textbf{\humaneval}}  \\
        \midrule
        \multirow{3}{13mm}{\vicuna-7B \;\; vs \vicuna-68M}
        &\const-5 & 03:50 & 20:01 & 25:53 & 120.38 \\
        &\const-10 & 05:05 & 19:30 & \red{\underline{26:16}} & \red{126.48} \\
        &\const-20 & 06:31 & \underline{19:25} & \blue{27:23} & \blue{\underline{125.38}} \\
        \midrule
        \multicolumn{6}{c}{\textbf{\gsm}}  \\
        \midrule
        \multirow{3}{13mm}{\vicuna-13B vs \vicuna-68M}
        &\const-5 & 07:03 & 45:16 & 1:12:54 & 365.47 \\
        &\const-10 & 09:14 & 44:02 & \red{\underline{1:11:55}} & \red{360.49} \\
        &\const-20 & 11:09 & \underline{40:45} & \blue{1:12:46} & \blue{\underline{357.97}} \\
        
        \midrule
        
        \multicolumn{6}{c}{\textbf{\cnndm}}  \\
        \midrule
        \multirow{3}{13mm}{\vicuna-13B vs \vicuna-68M}
        &\const-5 & 03:18 & 25:00 & 42:27 & 212.63 \\
        &\const-10 & 03:45 & 23:33 & \red{\underline{41:11}} & \red{202.32} \\
        &\const-20 & 04:12 & \underline{23:11} & \blue{41:20} & \blue{\underline{196.58}} \\
        \bottomrule
    \end{tabular}
    \vspace{-1mm}
    \caption{Assistant and target model run time in \const on \vicuna models. We see instances where a setup with lesser total time (colored in \red{red}) ends up in higher total energy consumption than another setup with higher total time (colored in \blue{blue}). 
    }
    \vspace{-6mm}
    \label{table:specdec-assistant-target-time}
\end{table}

 \begin{figure*}[!ht]
    \centering
    {\includegraphics[width=0.90\linewidth]{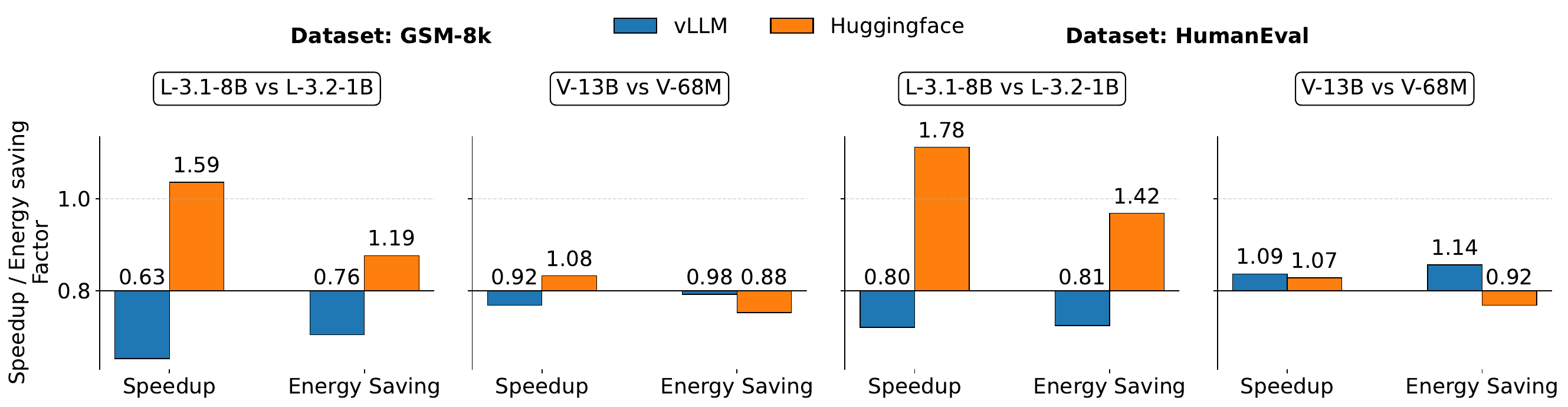}}%
    \caption{Comparison of speedup ($\gamma_t$) and energy savings ($\gamma_e^{Total}$) achieved by speculative decoding across different implementational platforms -- HuggingFace and vLLM %, using two assistant–target model pairs, 
    evaluated on the \gsm and \humaneval.  HuggingFace consistently outperforms vLLM in both metrics across almost all configurations.}
    \vspace{-4mm}
\label{fig:vllm}
\end{figure*}
For \humaneval, methods with speedup around $1\times$ (\const and \heuristic) generally translates in equivalent energy consumption with the vanilla (target model only) setup. 
However, for \gsm, methods with even higher speedup (around $1.25\times$) turns out to consume higher energy than \vanilla setup, again showing the dataset to be a influential factor deciding the interplay between energy optimization and walltime speedup. 

\subsection{Effect of Individual Model Runtime}

In this section we separately analyze runtime and energy contributions of the target and assistant models. Intuitively, the much larger target model accounts for the majority of the total energy consumption in the SD setup. Thus, we examine how the runtimes of both models influence time and energy.

Table~\ref{table:specdec-assistant-target-time} presents the assistant run time, target run time, and total run time separately. The total energy consumption of both models is also shown.We see several instances where a setup having \textit{lower total time} consumes \textit{higher total energy} compared to another setup. For instance, \const-20 setup often takes more total time but lower total energy than \const-10 setup, primarily due to its reduced target run time (underlined in Table~\ref{table:specdec-assistant-target-time}). 

\noindent \textbf{Overhead.} We also notice that an important factor that influences energy saving in SD is the \textit{presence of overheads}, i.e., additional computation and orchestration costs arising beyond direct token generation. 
These overheads primarily involve CPU and memory activity (i.e. managing cache/memory transfers, token verification, control flow logic), that, unlike GPU-bound computations, yield lower throughput and contribute disproportionately to energy consumption.
As a result, configurations that offer marginal speedups may end up with higher energy usage than vanilla decoding.
Our analysis reveals that these overheads can contribute approximately 6–10\% of total runtime in certain configurations. For instance, in the \humaneval dataset with \vicuna-7B and \const-10, the combined assistant and target runtimes sum to $\approx$24.5 minutes, while the total measured time is $\approx$26.2 minutes, indicating $\approx$1.7 minutes of overheads.

\subsection{Trends across Implementation Platforms}

Figure~\ref{fig:vllm} reports the runtime speedup and energy savings obtained via SD on HuggingFace and vLLM backends.
Across all configurations, HuggingFace consistently outperforms vLLM in both metrics.

This disparity is primarily attributable to differences in quantization strategy: HuggingFace employs NF4 4-bit quantization for target models, whereas vLLM uses GPTQ 4-bit quantization. The superior performance of HuggingFace suggests that NF4 better preserves SD efficiency by lowering verification overhead and improving compute–memory trade-offs. In contrast, GPTQ exacerbates speculative overheads, limiting achievable speedup and energy savings, indicating that the sustainability gains of speculative decoding are sensitive to backend design.

\vspace{-1mm}
\section{Concluding Discussions}
\vspace{-1mm}
This study is the first attempt towards benchmarking energy-consumption of various SD strategies under diverse model and task settings. Our primary takeaways are as follows: 
(1)~Lower inference time does not always correlate with proportional energy savings. Vanilla autoregressive decoding, despite higher latency, can sometimes be more energy-efficient than SD approaches (\const and \heuristic). 
(2)~Larger target-assistant model size gap generally results in better energy optimization.
(3)~Dataset characteristics play a critical role in energy reduction.
(4)~Correlation between runtime speedup and energy savings is affected by both the model architecture and decoding strategy; e.g., \llama models are more suited for energy savings than \vicuna models.

In summary, our findings highlight that speculative decoding is a promising approach for energy-efficient inference. But it is not a silver bullet, and we conclude by urging the community to consider the several factors needed to achieve energy-efficient inference in SD setup.

\section*{Limitations}

While our study provides valuable insights into the energy characteristics of speculative decoding strategies, several limitations must be acknowledged:
\begin{itemize}
    \item \textbf{Hardware Constraints:} All experiments were conducted on a fixed set of hardware configurations - specifically, NVIDIA A5000 and A6000 GPUs in a controlled local server environment. While this ensured consistency across measurements, it limits the generalizability of our results to other deployment environments, such as edge devices, multi-GPU clusters, or power-optimized cloud instances.

    \item \textbf{Scope of batch size:} Our evaluation primarily focuses on SD strategies with a batch size of one, a setting commonly adopted in academic research. However, in real-world deployment scenarios, the impact of larger batch sizes becomes critical, which may limit the generalizability of our findings.
    
\end{itemize}
Future work addressing these limitations can further refine our understanding of energy-efficient inference and contribute to the development of more generalizable and sustainable LLM deployment strategies.

\section*{Ethical Considerations}

One of the main ethical issues was the substantial energy consumption and carbon emissions generated by our experimente. 
We performed inferences over 3 datasets in several speculative decoding configurations, necessitating multiple repetitions of the inferences, along with several pilot experiments to finalize the experimental setup. This led to an approx total energy consumption of 1000 kWh. 
To reduce our environmental impact, we limited our experiments to only 256 test examples sampled from the datasets. We hope that the insights from this study will lead the community towards a much larger reduction of the energy consuption of LLMs.

\bibliography{custom}

\newpage
\noindent {\bf \Large Appendix}

\appendix
\section{Sample prompts}
\label{app:datasets}
For sample prompts from each dataset, refer to Table~\ref{table:dataset-gsm8k},~\ref{table:dataset-humaneval} and~\ref{table:dataset-cnndm}.

\section{Model configurations}
\label{app:model-details}
Refer to Table~\ref{tab:model-config} for models used and their aliases. 
\begin{table}[H]
    \scriptsize
    \centering
    \begin{tabular}{P{46mm}|P{29mm}}
        \toprule
        \textbf{Model} &
        \textbf{Alias} \\
        \midrule
        double7/vicuna-68m & \vicuna-68M / \vv-68M \\
        \midrule
        lmsys/vicuna-7b-v1.3 & \vicuna-7B / \vv-7B\\
        \midrule
        lmsys/vicuna-13b-v1.3 & \vicuna-13B / \vv-13B\\
        \midrule
        meta-llama/Llama-3.2-1B-Instruct & \llama-1B \\
        \midrule
        meta-llama/Llama-3.1-8B-Instruct & \llama-8B\\
        \midrule
        meta-llama/Llama-3.3-70B-Instruct & \llama-70B\\
        \midrule
         google/flan-t5-base & \flan-B\\
         \midrule
         google/flan-t5-large & \flan-L\\
         \midrule
         google/flan-t5-xl & \flan-XL\\
         \midrule
         Qwen/Qwen3-0.6B & \qwen-0.6B / \qw-0.6B\\
         \midrule
         Qwen/Qwen3-4B & \qwen-4B / \qw-4B\\
         \midrule
         Qwen/Qwen3-8B & \qwen-8B / \qw-8B\\
        \bottomrule
    \end{tabular}
    \caption{Model names and their alias}
    \label{tab:model-config}
\end{table}

\section{Speculative Decoding strategies}
\label{app:specdec-strat}
In the speculative decoding framework, \const-$x$ and \heuristic-$x$ are widely used variations for assistant generation. 
In \const-$x$, the assistant model generates $x$ number of tokens in each iteration. 
Whereas, in \heuristic-$x$, the assistant model generates $x$ tokens initially. After the target verification phase, if all the assistant tokens are accepted (by the target model), then in the next iteration, the assistant model generates $x+2$ tokens; otherwise, $x-1$ tokens are generated in the next iteration.
For \const-$x$, the value of $x$ was set to $5, 10$ and $20$ in our experiments.

\noindent 
\textbf{\medusa}~\cite{cai2023medusa} introduces an alternative design to mitigate the sequential bottleneck of autoregressive decoding. Instead of relying on a separate draft model, \medusa augments the target model with multiple decoding heads that predict several future tokens in parallel. These predictions are verified simultaneously using a tree-based attention mechanism, substantially reducing inference latency. The framework provides two fine-tuning modes: \medusa-1, which fine-tunes only the added heads for lossless acceleration, and \medusa-2, which jointly fine-tunes both the heads and the backbone for higher gains. By incorporating a typical acceptance scheme and an optional self-distillation procedure for data generation, \medusa achieves $2.3-2.8\times$ speedups on models like Vicuna and Zephyr with negligible quality degradation - demonstrating its simplicity, adaptability, and effectiveness in modern LLM systems.

\noindent \textbf{\eagle-2}~\cite{eagle2} builds on standard speculative decoding by introducing a context-aware dynamic draft tree for generating dynamic drafts. 
, which adapts draft token generation based on the confidence scores of a smaller draft model. This approach allows EAGLE-2 to increase the number of accepted tokens per cycle, leading to faster, lossless inference. Unlike prior methods using fixed tree structures (e.g., EAGLE \cite{eagle}, Medusa \cite{cai2023medusa} ), EAGLE-2 dynamically adjusts the tree shape without requiring additional training, achieving state-of-the-art speedups while preserving the original LLM's output distribution.

\noindent \textbf{\eagle-3}~\cite{eagle3} further eliminates the feature prediction constraint used in earlier versions like EAGLE and EAGLE-2. It introduces direct token prediction, allowing the draft model to integrate multi-layer fused features from the target model, significantly enhancing its expressiveness and scalability. 

\section{Hardware and Energy metrics}
\label{app:hardware}
We use the popular Code  Carbon~\cite{schmidt2021codecarbon} package to measure the energy consumed in different experiments. \citet{jay2023experimental} and \citet{bouza2023estimate} demonstrated the suitability and accuracy of CodeCarbon across various software-based power meter setups. This package measure the GPU-power usage using pynvml and CPU-power using Intel RAPL files every $\mathcal{X}$ seconds, and integrates it over time, which we set as $1secs$. Code-carbon also adds an estimate of the RAM-power being used depending on the RAM size. Power Usage Effectiveness~(PUE) is set to $1.0$ as all experiments are performed on the same server, indicating the actual energy usage may be different than reported. During inference, we provide test samples sequentially at batch size $1$ to the LLM and report the average energy usage per $1K$ tokens in Watt-hour (Wh). 

In our study, we omit the amount of carbon emission because we perform all the experiments in a single region where the carbon intensity is fixed and therefore, energy consumed is closely related with the amount of $CO_2$ emission. Furthermore, the CO$_2$ emission strongly varies depending on the region and the type of electricity source. Thus, we prefer to report the total energy consumed instead of the amount of CO$_2$ emission.

\section{Additional metrics}
\label{app:metrics}
We employ the following additional metrics to evaluate speculative decoding approaches in Table~\ref{table:specdec-energy-comp}:
\begin{itemize}
    \item \textbf{Total energy per $1K$ tokens:} The total energy consumed by target and assistant model measured in Watt (Wh) to generate $1K$ tokens. 

    \item \textbf{Total time per $1K$ tokens:} Total time in minutes required by both to generate $1K$ tokens. 
\end{itemize}

\section{Extended Results of \texorpdfstring{\const-$\mathcal{X}$}{const-X}}
\label{app:coga}

Table \ref{tab:coga} presents an extended evaluation of the Constant Generation by Assistant (COGA-x) strategy by varying the draft length 
$x \in {5, 10, 20}$ across all target–assistant model pairs and datasets. These results provide deeper insight into how the choice of draft length influences runtime speedup ($\gamma_t$) and energy savings ($\gamma_e$).
\\

\noindent\textbf{Effect of draft length:} Across most model families, increasing the draft length generally improves speedup, as larger drafts allow the target model to verify more tokens per iteration. However, this improvement does not always translate into proportional energy savings. In several cases, particularly for \vicuna-7B and \vicuna-13B, changes in $x$ yield only marginal variation in $\gamma_e^{GPU/Energy}$, with values often remaining close to or below unity despite moderate gains in $\gamma_t$. This suggests that verification and orchestration overheads dominate the energy profile for these models, limiting the benefit of longer drafts.
\\

\noindent\textbf{Model-family trends:} \llama-8B exhibits a consistent increase in both speedup and energy savings as $x$ increases, achieving up to $1.42\times$ total energy savings on \humaneval with \heuristic-20. In contrast, \llama-70B shows modest gains in speedup but persistent energy degradation on CNN-DM for all values of $x$, indicating that larger target models are more sensitive to dataset characteristics and overhead costs. Encoder-decoder models (\flan-L and \flan-XL) benefit most from larger drafts, where \const-20 consistently achieves the highest speedup and energy savings across datasets.
\\

\noindent\textbf{Dataset sensitivity:} The impact of $x$ is strongly dataset-dependent. \humaneval generally benefits from larger drafts, while \gsm and \cnndm often exhibit diminishing or negative energy returns as $x$ increases. This highlights that longer drafts can increase assistant computation and verification overhead without sufficient token acceptance to offset the added energy cost.
\\

\noindent Overall, Table \ref{tab:coga} demonstrates that while increasing the \const draft length can improve runtime speedup, energy efficiency gains are neither monotonic nor guaranteed. Optimal values of $x$ depend jointly on the target–assistant model pair and the dataset, reinforcing the need for adaptive or dataset-aware draft-length selection when energy efficiency is a primary objective.

\section{Acknowledgements}
The authors thank the anonymous reviewers whose suggestions helped to improve the work. The research is partially supported by a research grant from Accenture Corporation. Paramita Koley is supported by SERB NPDF Fellowship.

\begin{table*}[!ht]
    \scriptsize
    \centering
    \begin{tabular}{P{25mm} | P{120mm} }
    \toprule 
    \multicolumn{2}{c}{\textbf{\gsm}}\\
    \toprule 
     \textbf{model} & \textbf{prompt} \\    
    \midrule 
    \llama-3 & 
    <|begin\_of\_text|><|start\_header\_id|>system<|end\_header\_id|>

    Cutting Knowledge Date: December 2023
    Today Date: 11 May 2025
    
    You are a helpful, respectful and honest assistant. Always answer as helpfully as possible, while being safe. Your answers should not include any harmful, unethical, racist, sexist, toxic, dangerous, or illegal content. Please ensure that your responses are socially unbiased and positive in nature.
    
    If a question does not make any sense, or is not factually coherent, explain why instead of answering something not correct. If you don't know the answer to a question, please don't share false information.<|eot\_id|><|start\_header\_id|>user<|end\_header\_id|>
    
    Solve the math problem and give a numeric solution
    Problem: Carol and Jennifer are sisters from Los Angeles who love collecting signatures from celebrities. During their summer break from school, the sisters spend every afternoon collecting signatures. After five weeks, Carol and Jennifer compare their autograph books, counting up the number of signatures each sister has collected. Carol has 20 signatures in her book, and Jennifer has 44. The sisters have three more weeks of summer vacation, and they decide they want to reach 100 signatures between them by the end of the summer. How many signatures do the sisters need to collect to reach their goal?<|eot\_id|><|start\_header\_id|>assistant<|end\_header\_id|> \\
    \midrule 
    \vicuna & 
    A chat between a curious user and an artificial intelligence assistant. The assistant gives helpful, detailed, and polite answers to the user's questions. USER: Solve the math problem and give a numeric solution
    Problem: Carol and Jennifer are sisters from Los Angeles who love collecting signatures from celebrities. During their summer break from school, the sisters spend every afternoon collecting signatures. After five weeks, Carol and Jennifer compare their autograph books, counting up the number of signatures each sister has collected. Carol has 20 signatures in her book, and Jennifer has 44. The sisters have three more weeks of summer vacation, and they decide they want to reach 100 signatures between them by the end of the summer. How many signatures do the sisters need to collect to reach their goal? ASSISTANT: \\
    \midrule
    \qwen & $<|im\_start|>$user
    You are a helpful, respectful and honest assistant. Always answer as helpfully as possible, while being safe. Your answers should not include any harmful, unethical, racist, sexist, toxic, dangerous, or illegal content. Please ensure that your responses are socially unbiased and positive in nature.
    
    If a question does not make any sense, or is not factually coherent, explain why instead of answering something not correct. If you don't know the answer to a question, please don't share false information.
    Solve the math problem and give a numeric solution
    Problem: Carol and Jennifer are sisters from Los Angeles who love collecting signatures from celebrities. During their summer break from school, the sisters spend every afternoon collecting signatures. After five weeks, Carol and Jennifer compare their autograph books, counting up the number of signatures each sister has collected. Carol has 20 signatures in her book, and Jennifer has 44. The sisters have three more weeks of summer vacation, and they decide they want to reach 100 signatures between them by the end of the summer. How many signatures do the sisters need to collect to reach their goal?$<|im\_end|>$
    $<|im\_start|>$assistant
    \\
    \bottomrule
    \end{tabular}
    \caption{Prompts for the \gsm dataset}
    \label{table:dataset-gsm8k}
\end{table*}

\begin{table*}[!ht]
    \scriptsize
    \centering
    \begin{tabular}{P{25mm} | P{120mm} }
    \toprule 
    \multicolumn{2}{c}{\textbf{\humaneval}}\\
    \toprule 
     \textbf{model} & \textbf{prompt} \\    
    \midrule

    \llama-3 & <|begin\_of\_text|><|start\_header\_id|>system<|end\_header\_id|>

    Cutting Knowledge Date: December 2023
    Today Date: 10 May 2025
    
    You are a helpful, respectful and honest assistant. Always answer as helpfully as possible, while being safe. Your answers should not include any harmful, unethical, racist, sexist, toxic, dangerous, or illegal content. Please ensure that your responses are socially unbiased and positive in nature.
    
    If a question does not make any sense, or is not factually coherent, explain why instead of answering something not correct. If you don't know the answer to a question, please don't share false information.<|eot\_id|><|start\_header\_id|>user<|end\_header\_id|>
    
    Complete the function(s) based on the given function prototype and the docstring:
    
    def can\_arrange(arr):
        """"""Create a function which returns the largest index of an element which
        is not greater than or equal to the element immediately preceding it. If
        no such element exists then return -1. The given array will not contain
        duplicate values.
    
        Examples:
        can\_arrange([1,2,4,3,5]) = 3
        can\_arrange([1,2,3]) = -1
        """"""<|eot\_id|><|start\_header\_id|>assistant<|end\_header\_id|>
    \\
    \midrule 
    \vicuna & A chat between a curious user and an artificial intelligence assistant. The assistant gives helpful, detailed, and polite answers to the user's questions. USER: Complete the function(s) based on the given function prototype and the docstring:

    def can\_arrange(arr):
        """"""Create a function which returns the largest index of an element which
        is not greater than or equal to the element immediately preceding it. If
        no such element exists then return -1. The given array will not contain
        duplicate values.
    
        Examples:
        can\_arrange([1,2,4,3,5]) = 3
        can\_arrange([1,2,3]) = -1
        """""" ASSISTANT:
    \\
    \midrule
    \qwen & $<|im\_start|>$user
    You are a helpful, respectful and honest assistant. Always answer as helpfully as possible, while being safe. Your answers should not include any harmful, unethical, racist, sexist, toxic, dangerous, or illegal content. Please ensure that your responses are socially unbiased and positive in nature.
    
    If a question does not make any sense, or is not factually coherent, explain why instead of answering something not correct. If you don't know the answer to a question, please don't share false information.
    Complete the function(s) based on the given function prototype and the docstring:
    
    def can\_arrange(arr):
        """"""Create a function which returns the largest index of an element which
        is not greater than or equal to the element immediately preceding it. If
        no such element exists then return -1. The given array will not contain
        duplicate values.
    
        Examples:
        can\_arrange([1,2,4,3,5]) = 3
        can\_arrange([1,2,3]) = -1
        """"""$<|im\_end|>$
    $<|im\_start|>$assistant
    \\
    \bottomrule
    \end{tabular}
    \caption{Prompts for the \humaneval dataset}
    \label{table:dataset-humaneval}
\end{table*}

\begin{table*}[!ht]
    %\footnotesize
    \tiny
    \centering
    \begin{tabular}{P{10mm} | P{140mm} }
    \toprule 
    \multicolumn{2}{c}{\textbf{\cnndm}}\\
    \toprule 
     \textbf{model} & \textbf{prompt} \\    
    \midrule 
    \llama-3 & <|begin\_of\_text|><|start\_header\_id|>system<|end\_header\_id|>
    
    Cutting Knowledge Date: December 2023
    Today Date: 11 May 2025
    
    You are a helpful, respectful and honest assistant. Always answer as helpfully as possible, while being safe. Your answers should not include any harmful, unethical, racist, sexist, toxic, dangerous, or illegal content. Please ensure that your responses are socially unbiased and positive in nature.
    
    If a question does not make any sense, or is not factually coherent, explain why instead of answering something not correct. If you don't know the answer to a question, please don't share false information.<|eot\_id|><|start\_header\_id|>user<|end\_header\_id|>
    
    Summarize the following news article in about 50 words: ARTICLE: Down Augusta way they say the azaleas are in full bloom, which is more than can be said for England’s Justin Rose. A bruising Florida swing last month saw the Englishman fall outside the world’s top 10. For a player who has been virtually a fixture in the top five for the last three years it was certainly a dent to the ego, with the Masters now just around the corner. Rose’s solution to his miserable form — three missed cuts and a 55th-place finish at the Cadillac Championship in four PGA Tour starts — was the time-honoured one. For the past two weeks, the 34-year-old has spent long hours on the practice ground. Justin Rose hit 17 out of 18 greens in regulation and signed for a 69 at the Shell Houston Open . In the first round of the Shell Houston Open on Thursday there were encouraging signs his decline will prove temporary. Rose hit 17 out of 18 greens in regulation and signed for a 69, the same score as his playing partner, the ever- consistent Jordan Spieth. ‘It’s certainly a welcome return to the sixties, for it had been a while,’ said Rose, smiling. On a day when American Scott Piercy went round in 63 and Phil Mickelson enjoyed his best round in months with a 66, it was hardly surprising the only reporter waiting to talk to Rose was this one. But under the radar is never a bad place to be going to the Masters. The boom and bust years that characterised the first half of Rose’s career meant there was never going to be any feelings of panic following his unusually poor run in the Sunshine State. ‘There’s no doubt I lost my game there but the Florida swing can be unforgiving if you’re slightly off,’ he said. ‘Over the past two weeks I feel like I’ve done some good work and whether I finish well or not here I feel like I’m going in the right direction again. ‘Basically I was getting ahead of the ball at impact, and shots were going left or right, the irons were not solid and the new putter was not working. So we’ve corrected the faults and I’ve gone back to the old putter.’ Phil Mickelson enjoyed his best round in months with a 66 on Thursday . Does he pay much attention to the world rankings? ‘You notice, for sure,’ he said. ‘I’m very proud of the fact I’ve been in the world’s top five for practically the whole of the last three years. It’s a nice ego thing, so by the end of the year I’m hoping there won’t be any slippage. ‘But right now, I’ve got to focus on my game in the knowledge that the rankings change fast when you’re playing well. ...... ‘The Masters has probably been less on my mind this year because I am trying to find some form,’ he admitted. ‘But I think the fact I’ve had a number of great rounds there will always stand me in good stead. Regardless of what happens here, I feel comfortable on that course and know I can manage my game even if it’s not 100 per cent. You draw off the energy of the place.’ Mickelson has certainly done that over the years and perhaps the veteran lefty, a three-time Masters champion, is gearing himself up for another run at the green jacket. ‘It was a good start to the tournament and now I’m looking for three more good rounds,’ he said. 'This is a big week for me. I felt the game was close last week. The only thing missing was chipping and short game.' Paul Casey, like Mickelson another former winner of this event, celebrated his last-gasp Masters invitation with a fine round of 68 notable for two eagle threes. In the afternoon wave, Padraig Harrington and Lee Westwood both made good starts to play their first six holes in two under.<|eot\_id|><|start\_header\_id|>assistant<|end\_header\_id|>
    \\
    \midrule 
    \vicuna & A chat between a curious user and an artificial intelligence assistant. The assistant gives helpful, detailed, and polite answers to the user's questions. USER: Summarize the following news article in about 50 words: ARTICLE: Down Augusta way they say the azaleas are in full bloom, which is more than can be said for England’s Justin Rose. A bruising Florida swing last month saw the Englishman fall outside the world’s top 10. For a player who has been virtually a fixture in the top five for the last three years it was certainly a dent to the ego, with the Masters now just around the corner. Rose’s solution to his miserable form — three missed cuts and a 55th-place finish at the Cadillac Championship in four PGA Tour starts — was the time-honoured one. For the past two weeks, the 34-year-old has spent long hours on the practice ground. Justin Rose hit 17 out of 18 greens in regulation and signed for a 69 at the Shell Houston Open . In the first round of the Shell Houston Open on Thursday there were encouraging signs his decline will prove temporary. Rose hit 17 out of 18 greens in regulation and signed for a 69, the same score as his playing partner, the ever- consistent Jordan Spieth. ‘It’s certainly a welcome return to the sixties, for it had been a while,’ said Rose, smiling. On a day when American Scott Piercy went round in 63 and Phil Mickelson enjoyed his best round in months with a 66, it was hardly surprising the only reporter waiting to talk to Rose was this one. But under the radar is never a bad place to be going to the Masters. The boom and bust years that characterised the first half of Rose’s career meant there was never going to be any feelings of panic following his unusually poor run in the Sunshine State. ‘There’s no doubt I lost my game there but the Florida swing can be unforgiving if you’re slightly off,’ he said. ‘Over the past two weeks I feel like I’ve done some good work and whether I finish well or not here I feel like I’m going in the right direction again. ‘Basically I was getting ahead of the ball at impact, and shots were going left or right, the irons were not solid and the new putter was not working. So we’ve corrected the faults and I’ve gone back to the old putter.’ Phil Mickelson enjoyed his best round in months with a 66 on Thursday . Does he pay much attention to the world rankings? ‘You notice, for sure,’ he said. ‘I’m very proud of the fact I’ve been in the world’s top five for practically the whole of the last three years. It’s a nice ego thing, so by the end of the year I’m hoping there won’t be any slippage. ‘But right now, I’ve got to focus on my game in the knowledge that the rankings change fast when you’re playing well. .... . ‘The Masters has probably been less on my mind this year because I am trying to find some form,’ he admitted. ‘But I think the fact I’ve had a number of great rounds there will always stand me in good stead. Regardless of what happens here, I feel comfortable on that course and know I can manage my game even if it’s not 100 per cent. You draw off the energy of the place.’ Mickelson has certainly done that over the years and perhaps the veteran lefty, a three-time Masters champion, is gearing himself up for another run at the green jacket. ‘It was a good start to the tournament and now I’m looking for three more good rounds,’ he said. 'This is a big week for me. I felt the game was close last week. The only thing missing was chipping and short game.' Paul Casey, like Mickelson another former winner of this event, celebrated his last-gasp Masters invitation with a fine round of 68 notable for two eagle threes. In the afternoon wave, Padraig Harrington and Lee Westwood both made good starts to play their first six holes in two under. ASSISTANT:
    \\
    \midrule
    \qwen & $<|im\_start|>$ user You are a helpful, respectful and honest assistant. Always answer as helpfully as possible, while being safe. Your answers should not include any harmful, unethical, racist, sexist, toxic, dangerous, or illegal content. Please ensure that your responses are socially unbiased and positive in nature.
    If a question does not make any sense, or is not factually coherent, explain why instead of answering something not correct. If you don't know the answer to a question, please don't share false information. Summarize the following news article in about 50 words: ARTICLE: Down Augusta way they say the azaleas are in full bloom, which is more than can be said for England’s Justin Rose. A bruising Florida swing last month saw the Englishman fall outside the world’s top 10. For a player who has been virtually a fixture in the top five for the last three years it was certainly a dent to the ego, with the Masters now just around the corner. Rose’s solution to his miserable form — three missed cuts and a 55th-place finish at the Cadillac Championship in four PGA Tour starts — was the time-honoured one. For the past two weeks, the 34-year-old has spent long hours on the practice ground. Justin Rose hit 17 out of 18 greens in regulation and signed for a 69 at the Shell Houston Open . In the first round of the Shell Houston Open on Thursday there were encouraging signs his decline will prove temporary. Rose hit 17 out of 18 greens in regulation and signed for a 69, the same score as his playing partner, the ever- consistent Jordan Spieth. ‘It’s certainly a welcome return to the sixties, for it had been a while,’ said Rose, smiling. On a day when American Scott Piercy went round in 63 and Phil Mickelson enjoyed his best round in months with a 66, it was hardly surprising the only reporter waiting to talk to Rose was this one. But under the radar is never a bad place to be going to the Masters. The boom and bust years that characterised the first half of Rose’s career meant there was never going to be any feelings of panic following his unusually poor run in the Sunshine State. ‘There’s no doubt I lost my game there but the Florida swing can be unforgiving if you’re slightly off,’ he said. ‘Over the past two weeks I feel like I’ve done some good work and whether I finish well or not here I feel like I’m going in the right direction again. ‘Basically I was getting ahead of the ball at impact, and shots were going left or right, the irons were not solid and the new putter was not working. So we’ve corrected the faults and I’ve gone back to the old putter.’ Phil Mickelson enjoyed his best round in months with a 66 on Thursday . Does he pay much attention to the world rankings? ‘You notice, for sure,’ he said. ‘I’m very proud of the fact I’ve been in the world’s top five for practically the whole of the last three years. It’s a nice ego thing, so by the end of the year I’m hoping there won’t be any slippage. ‘But right now, I’ve got to focus on my game in the knowledge that the rankings change fast when you’re playing well. ........ ‘The Masters has probably been less on my mind this year because I am trying to find some form,’ he admitted. ‘But I think the fact I’ve had a number of great rounds there will always stand me in good stead. Regardless of what happens here, I feel comfortable on that course and know I can manage my game even if it’s not 100 per cent. You draw off the energy of the place.’ Mickelson has certainly done that over the years and perhaps the veteran lefty, a three-time Masters champion, is gearing himself up for another run at the green jacket. ‘It was a good start to the tournament and now I’m looking for three more good rounds,’ he said. 'This is a big week for me. I felt the game was close last week. The only thing missing was chipping and short game.' Paul Casey, like Mickelson another former winner of this event, celebrated his last-gasp Masters invitation with a fine round of 68 notable for two eagle threes. In the afternoon wave, Padraig Harrington and Lee Westwood both made good starts to play their first six holes in two under.$<|im\_end|>$
    $<|im\_start|>$assistant
    \\
    \bottomrule
    \end{tabular}
    \caption{Prompts for the \cnndm dataset}
    \label{table:dataset-cnndm}
\end{table*}

\clearpage
\onecolumn

\begin{table*}[t]
  \centering
  \scriptsize
  \begin{tabular}{
        p{17mm} p{11mm}| 
        p{9mm} p{10mm} p{10mm}| 
        p{9mm} p{9mm} p{9mm}|
        p{9mm} p{9mm} p{9mm}}
        \toprule
        &  
        & \multicolumn{3}{c|}{\textbf{\humaneval}} 
        & \multicolumn{3}{c|}{\textbf{\gsm}} 
        & \multicolumn{3}{c}{\textbf{\cnndm}}         
        \\        
        \midrule
        & &
        \multicolumn{1}{c}{\textbf{Speedup}} &
        \multicolumn{2}{c|}{\textbf{Energy Saving Factor}} &
        \multicolumn{1}{c}{\textbf{Speedup}} &
        \multicolumn{2}{c|}{\textbf{Energy Saving Factor}} &
        \multicolumn{1}{c}{\textbf{Speedup}} &
        \multicolumn{2}{c}{\textbf{Energy Saving Factor}} \\
        \cmidrule(lr){3-3} \cmidrule(lr){4-5}
        \cmidrule(lr){6-6} \cmidrule(lr){7-8}
        \cmidrule(lr){9-9} \cmidrule(lr){10-11}
        
        \makecell{\textbf{Target vs}\\\textbf{Assistant}} &
        \makecell{\textbf{SD}\\\textbf{Method}} &
        $\boldsymbol{\displaystyle\gamma_t}$ &
        \makecell{$\boldsymbol{\displaystyle\gamma_e^{GPU}}$} &
        \makecell{$\boldsymbol{\displaystyle\gamma_e^{Total}}$} &
        $\boldsymbol{\displaystyle\gamma_t}$ &
        \makecell{$\boldsymbol{\displaystyle\gamma_e^{GPU}}$} &
        \makecell{$\boldsymbol{\displaystyle\gamma_e^{Total}}$} &
        $\boldsymbol{\displaystyle\gamma_t}$ &
        \makecell{$\boldsymbol{\displaystyle\gamma_e^{GPU}}$} &
        \makecell{$\boldsymbol{\displaystyle\gamma_e^{Total}}$} 
        \\
        \midrule
        \multirow{3}{*}{\makecell[l]{\vicuna-7B vs\\\vicuna-68M}}  
        & \const-5 & $1.44\times$ & $0.81\times$ & $0.97\times$ & $1.31\times$ & $0.75\times$ & $0.85\times$ & $1.38\times$ & $0.84\times$ & $0.96\times$ \\
        &  \const-10 & $1.45\times$ & $0.83\times$ & $0.99\times$ & $1.28\times$ & $0.75\times$ & $0.81\times$ & $1.34\times$ & $0.83\times$ & $0.95\times$ \\
        & \const-20 & $1.42\times$ & $0.83\times$ & $0.99\times$ & $1.19\times$ & $0.72\times$ & $0.77\times$ & $1.44\times$ & $0.89\times$ & $1.03\times$ \\
        \midrule
        \multirow{3}{*}{\makecell[l]{\vicuna-13B vs \\ \vicuna-68M}} & 
        \const-5 & $1.05\times$ & $0.89\times$ & $0.91\times$ & $1.08\times$ & $0.73\times$ & $0.85\times$ & $0.99\times$ & $0.86\times$ & $0.92\times$ \\
        & \const-10 & $1.10\times$ & $0.95\times$ & $0.93\times$ & $1.11\times$ & $0.76\times$ & $0.89\times$ & $1.02\times$ & $0.89\times$ & $0.96\times$ \\
        & \const-20 & $1.07\times$ & $0.93\times$ & $0.92\times$ & $1.08\times$ & $0.75\times$ & $0.88\times$ & $1.02\times$ & $0.89\times$ & $0.96\times$ \\
        \midrule
        \multirow{3}{*}{\makecell[l]{\llama-8B vs \\ \llama-1B}} 
        & \const-5 & $1.59\times$ & $1.06\times$ & $1.25\times$ & $1.42\times$ & $0.96\times$ & $1.01\times$ & $1.10\times$ & $0.86\times$ & $0.92\times$ \\
        & \const-10 & $1.75\times$ & $1.20\times$ & $1.38\times$ & $1.56\times$ & $1.06\times$ & $1.11\times$ & $1.05\times$ & $0.83\times$ & $0.89\times$ \\
        & \const-20 & $1.78\times$ & $1.23\times$ & $1.42\times$ & $1.59\times$ & $1.10\times$ & $1.19\times$ & $0.96\times$ & $0.79\times$ & $0.83\times$ \\
        \midrule
        \multirow{3}{*}{\makecell[l]{\llama-70B vs \\ \llama-1B}} 
        & \const-5 & $0.92\times$ & $0.94\times$ & $0.95\times$ & $0.85\times$ & $0.87\times$ & $0.88\times$ & $0.62\times$ & $0.64\times$ & $0.63\times$ \\
        & \const-10 & $1.13\times$ & $1.18\times$ & $1.17\times$ & $1.02\times$ & $1.04\times$ & $1.06\times$ & $0.63\times$ & $0.65\times$ & $0.65\times$ \\
        & \const-20 & $1.25\times$ & $1.33\times$ & $1.31\times$ & $1.09\times$ & $1.12\times$ & $1.14\times$ & $0.61\times$ & $0.64\times$ & $0.63\times$ \\
        \midrule
        \multirow{3}{*}{\makecell[l]{\flan-L vs \\ \flan-B}}
        & \const-5 & $1.73\times$ & $1.74\times$ & $1.70\times$ & $1.32\times$ & $1.32\times$ & $1.32\times$ & $1.38\times$ & $1.32\times$ & $1.32\times$\\
        & \const-10 & $1.90\times$ & $1.79\times$ & $1.80\times$ & $1.38\times$ & $1.36\times$ & $1.38\times$ & $1.51\times$ & $1.40\times$ & $1.42\times$\\
        & \const-20 & $2.01\times$ & $2.02\times$ & $2.00\times$ & $1.22\times$ & $1.22\times$ & $1.22\times$ & $1.47\times$ & $1.39\times$ & $1.40\times$\\
        \midrule
        \multirow{3}{*}{\makecell[l]{\flan-XL vs \\ \flan-B}}
        & \const-5 & $1.64\times$ & $1.62\times$ & $1.60\times$ & $1.06\times$ & $1.02\times$ & $1.01\times$ & $1.43\times$ & $1.32\times$ & $1.33\times$ \\
        & \const-10 & $1.71\times$ & $1.67\times$ & $1.70\times$ & $1.03\times$ & $1.01\times$ & $1.00\times$ & $1.43\times$ & $1.37\times$ & $1.38\times$\\
        & \const-20 & $1.86\times$ & $1.82\times$ & $1.81\times$ & $0.92\times$ & $0.92\times$ & $0.92\times$ & $1.69\times$ & $1.45\times$ & $1.45\times$\\
        \bottomrule
    \end{tabular}
    \caption{Comparative analysis of speedup ($\gamma_t$), GPU ($\displaystyle\gamma_e^{GPU}$) and Total ($\displaystyle\gamma_e^{Total}$) energy saving factor across multiple \const-$x$ speculative decoding strategies for varying values of $x \in {5, 10, 20}$.}
    \vspace{3mm}
    \label{tab:coga}

  \vspace{6pt}
          \begin{tabular}{P{12.5mm} P{14mm} |P{15mm} P{15mm}| P{15mm} P{15mm}| P{15mm} P{15mm}}
        \toprule
        &  
        & \multicolumn{2}{c|}{\textbf{\humaneval}} 
        & \multicolumn{2}{c|}{\textbf{\gsm}}
        & \multicolumn{2}{c}{\textbf{\cnndm}}         
         \\        
        \midrule
        \textbf{Target vs Draft} & \textbf{Method} & \textbf{GPU energy (Wh/1K tokens)} & \textbf{Total energy (Wh/1K tokens)} & \textbf{GPU energy (Wh/1K tokens)} & \textbf{Total energy (Wh/1K tokens)} & \textbf{GPU energy (Wh/1K tokens)} & \textbf{Total energy (Wh/1K tokens)}\\
        \midrule
        \multirow{5}{*}{\makecell[c]{\vicuna-7B \\ vs \\ \vicuna-68M}}  
        & \const-5 & 2.24 & 3.03 & 2.37 & 3.13 & 2.74 & 3.63 \\
        % \hline
        &  \const-10 & 2.19 & 2.98 & 2.37 & 3.30 & 2.76 & 3.65 \\
        % \hline
        & \const-20 & 2.19 & 2.98 & 2.46 & 3.46 & 2.59 & 3.38 \\
        % \hline
        & \heuristic-20 & 2.25 & 3.07 & 2.44 & 3.18 & 2.8 & 3.68 \\
        % \hline
        & \eagle2 & 1.15 & 1.55 & 1.30 & 1.66 & 1.97 & 2.51 \\
        \midrule
        \multirow{7}{*}{\makecell[c]{\vicuna-13B \\ vs \\ \vicuna-68M}} & 
        \const-5 & 3.83 & 5.02 & 4.00 & 5.27 & 4.55 & 6.12 \\
        % \hline
        & \const-10 & 3.60 & 4.87 & 3.85 & 5.02 & 4.39 & 5.9 \\
        % \hline
        & \const-20 & 3.65 & 4.93 & 3.90 & 5.07 & 4.38 & 5.88 \\
        % \hline
        & \heuristic-20 & 3.86 & 4.94 & 4.13 & 5.55 & 4.62 & 6.20 \\
        % \hline
        & \eagle2 & 1.91 & 2.49 & 2.01 & 2.75 & 3.04 & 3.85 \\
        % \hline
        & \eagle3 & 1.42 & 1.81 & 1.57 & 2.14 & 2.11 & 2.69 \\
        \midrule
        \multirow{6}{*}{\makecell[c]{\llama-8B \\ vs \\ \llama-1B}} 
        & \const-5 & 2.03 & 2.72 & 2.25 & 3.22 & 3.11 & 4.31 \\
        % \hline
        & \const-10 & 1.80 & 2.46 & 2.03 & 2.93 & 3.20 & 4.45 \\
        % \hline
        & \const-20 & 1.75 & 2.38 & 1.95 & 2.74 & 3.40 & 4.76 \\
        % \hline
        & \heuristic-20 & 1.71 & 2.35 & 2.02 & 2.92 & 3.34 & 4.65 \\
        % \hline
        & \eagle2 & 1.60 & 2.08 & 1.81 & 2.43 & 2.58 & 3.28 \\
        % \hline
        & \eagle3 & 1.24 & 1.62 & 1.36 & 1.88 & 2.03 & 2.60\\
        \midrule
        \multirow{5}{*}{\makecell[c]{\llama-70B \\ vs \\ \llama-1B}} 
        & \const-5 & 12.33 & 14.84 & 13.62 & 16.74 & 23.66 & 28.95 \\
        % \hline
        & \const-10 & 9.87 & 11.95 & 11.33 & 13.92 & 23.1 & 28.16 \\
        % \hline
        & \const-20 & 8.78 & 10.71 & 10.56 & 13.02 & 23.53 & 28.71 \\
        % \hline
        & \heuristic-20 & 8.44 & 10.29 & 10.36 & 12.92 & 23.34 & 28.53 \\
        % \hline
        & \eagle3 & 8.68 & 10.46 & 9.40 & 11.53 & 19.28 & 23.71\\
        \bottomrule
    \end{tabular}
    \caption{Comparative evaluation of inference energy consumption in various speculative decoding strategies for four standard target models on \humaneval, \gsm and \cnndm datasets. Energy is measured in Watt-Hour per $1000$ tokens (Wh/1K). In each case, corresponding draft model is mentioned. However, \eagle2 and \eagle3 employ draft model provided by the source repositories. }
    \label{table:specdec-energy-comp}

\end{table*}

\end{document}